\pdfoutput=1

\documentclass[11pt]{article}

\usepackage[final]{acl}

\usepackage{times}
\usepackage{latexsym}

\usepackage[T1]{fontenc}

\usepackage[utf8]{inputenc}

\usepackage{microtype}

\usepackage{inconsolata}

\usepackage{epstopdf}
\usepackage{booktabs}
\usepackage{multirow} 
\usepackage{makecell}
\usepackage{graphicx}
\usepackage{enumerate}
\usepackage{makecell}
\usepackage{supertabular}
\usepackage{amsmath}
\usepackage[linesnumbered,ruled,vlined]{algorithm2e}

\title{Prove Your Point!: Bringing Proof-Enhancement Principles to Argumentative Essay Generation}

\author{Ruiyu Xiao$^1$\quad
Lei Wu$^1$\quad
Yuhang Gou$^1$\quad
Weinan Zhang$^1$\quad
Ting Liu$^1$\footnotemark[2]\quad \\
$^1$\normalsize{Research Center for Social Computing and Information Retrieval}\\[-.05cm]
\normalsize{Harbin Institute of Technology, China}\\[-.05cm]
{\small\tt\{ryxiao, lwu, yhgou, wnzhang, tliu\}@ir.hit.edu.cn}\\
}

\begin{document}
\maketitle
\renewcommand{\thefootnote}{\fnsymbol{footnote}}
\footnotetext[2]{Corresponding author.}
\begin{abstract}
Argumentative essay generation (AEG) aims to generate complete texts on specific controversial topics or debates. 
Although current AEG methods can generate individual opinions, they often overlook the high-level connections between these opinions. This often leads to the generated results being mired in logical confusion, unable to proof their own arguments effectively.
The generated essay may present evidence that contradicts the claims or they may fail to assemble the claims into logical flow. 
In this paper, we present a unified two-stage framework: Proof-Enhancement and Self-Annotation (PESA) for AEG with a focus on logical enhancement.
Specifically, we first construct pseudo-labels for logical information,claims and grounds, using a large language model.
We then propose a tree planning approach that introduces proof principles and ensures logical consistency.
Extensive experimental results show that, benefiting from proof principle guidance, PESA generates argumentative essays with better logical validity and persuasiveness than strong baseline models.
\end{abstract}

\section{Introduction}

\begin{figure}[t]
\centering
\includegraphics[width=0.45\textwidth]{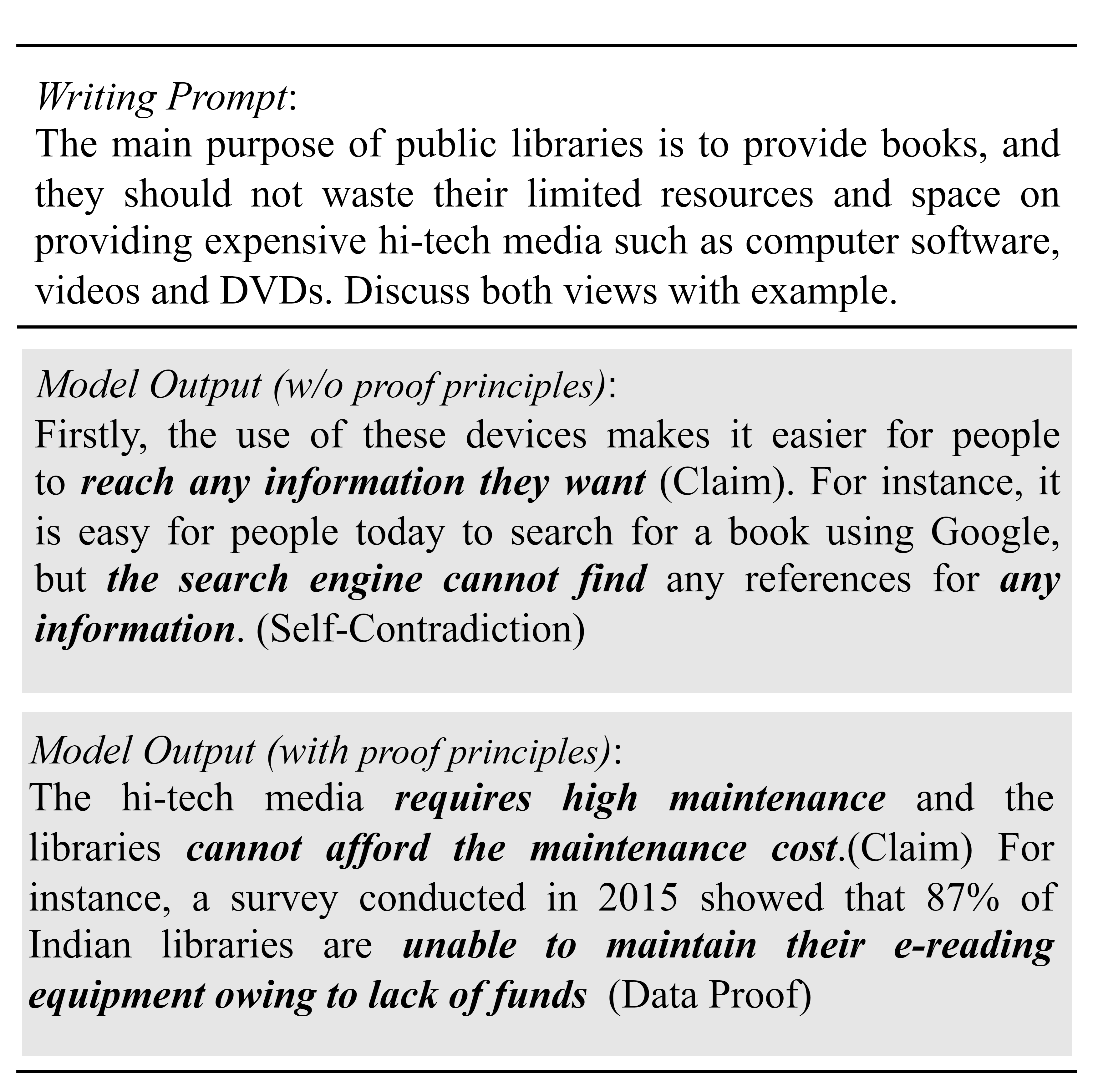}
\caption{Two examples of proof and logical disorganization leading to impaired persuasiveness. Obviously, the upper example gives self-contradiction claim and ground, the lower example gives correct and persuasive proof. }
\label{passage_example}
\end{figure}

From decision-making to public discussions, argumentative texts serve to proclaim ideas or defend a point of view in a wide range of scenarios \citet{sato-etal-2015-end}. Argumentative essay generation (AEG) is a task designed to generate a persuasive argumentative text containing several arguments on a given controversial debate topic (e.g., whether euthanasia should be legalized). \citet{bao-etal-2022-aeg} introduced a keywords text-planning method to the AEG task. The plan-and-write paradigm used in this method enriches the length and content of argumentative text, yielding impressive results. However, due to the lack of logical and proof guidance in planning, current AEG methods tend to overlook the high-level connections between individual opinions. 

Existing AEG methods, although capable of generating meaningful opinions, consistently struggle with providing compelling proof. Specifically, the generated essay may present evidence that contradicts the claims or they may fail to assemble the claims into logical flow. In Figure \ref{passage_example}, we give a example of logical disorganization leading to impaired persuasiveness. In the upper example of Figure \ref{passage_example}, we observe that the data and evidence given in the paragraph not only fail to proof the claim but even contradict it. These instances highlight a significant challenge in the AEG task: the generated claims and evidence are not not been proven properly and organize into logically consistent, persuasive argumentative texts.  

Human writers encounter difficulties in providing evidence when engaging in argumentative writing. To address this, educators have introduced a set of proof guidelines and norms to guide students on how to complete correct proofs \citep{lee2014argument}. For instance, \citet{toulmin2003uses} introduced the Toulmin Argumentation Model, positing that a complete and convincing proof should incorporate abstract arguments, which refer to the overarching claims or positions, and specific grounds, data, or warrants, which provide concrete evidence or justifications for these claims. 
These proof principles or norms provide students with clear guidance, requiring the use of a correct logical structure in argumentative essays, and a careful examination of the logical flow between claims and the evidences or data supporting each claim. Subsequent analysis by educators\citep{stapleton2015assessing,ozfidan2020detected} shows that this proof-enhancement instruction significantly improves the persuasiveness of students' argumentative writing.

Inspired by this, we introduce explicit proof principles and norms to guide the model in generating logical and persuasive argumentative texts named Proof-Enhancement planning (PE). Specifically, we refer to the widely adopted Toulmin Argumentation Model, using abstract claims and specific grounds as tree planning approach to generate accurate and complete proof processes. In the process of implementing PE, we discovered that acquiring high-quality proof-enhanced planning data necessitates extensive manual annotation by professionals, making it exceedingly difficult to obtain. Consequently, we introduced Self-Annotation (SA), which leverages prior knowledge and employs a layer-by-layer extraction approach to obtain pseudo-labels for proof information, thereby expanding the existing AEG dataset to address the lack of fine-grained logical annotation data.

Specifically, we propose a unified two-stage framework known as the Proof-Enhancement and Self-Annotation methods (PESA). As shown in Figure \ref{model-graph}, In the Proof-Enhancement phase, the correctness and completeness of the generated argumentative text are enhanced, addressing the issue of logical disorganization present in existing AEG methods. The Self-Annotation phase leverages prior knowledge to automatically extract fine-grained logical information, solving the problem of proof principles requiring extensive manual annotation by experts.

Extensive experimental results on AEG dataset show that PESA achieves State-Of-Art performance across various automatic evaluation metrics, especially on logical structure and persuasion. Meanwhile, human evaluation further confirms that PESA significantly outperforms the strong baseline with its superiority in terms of metrics such as fluency, logic and persuasiveness. Besides, we find that PESA achieves comparable results to ChatGPT and other LLMs by introducing proof principles and logical information, even when there is a large gap in parameter sizes.

To conclude, our model made the following contributions
\begin{itemize}
    \item  To the best of our knowledge, we are the first to consider proof principles and norms for the argumentative essay generation task.  
    \item We design a Self-Annotation method that automatically completes the fine-grained annotation of AEG, and constructs high-quality AEG data with logical information.
    \item We propose a Proof-Enhancement method that mimics the Toulmin Argumentation Model, enabling the model to generate a better proof and more persuasive argumentative texts.
    \item Our method demonstrates superior performance over existing models across a wide range of human and automatic evaluation metrics and achieves comparable results to ChatGPT with much smaller parameters.
\end{itemize}

\section{Related Work}
\subsection{Text-planning Method in Neural Generation}
Given that natural language generation models often produce texts that lack relevance to the topic or exhibits a deficiency in high-level structure\citep{zhao-etal-2017-learning,xu-etal-2020-discourse}, the idea of using external information as text-planning to guide generative models has a long history. 

Currently, researchers have shown great interest in applying text-planning to large-scale pre-trained language models such as GPT. \citet{tan-etal-2021-progressive} used several keywords as planning and progressively refined them via multiple stages into generate coherent long texts.  \citet{mirowski2023co} built structural context via prompt chaining and large language model to enhance usefulness for long-form creative writing. \citet{guan2023generating} extended Transformer model to dynamically learn and update text planning, which led a more coherent and diverse stories and news generation. 

The study of the AEG began relatively late and research on relevant methodologies is still in its infancy due to its remarkable challenging nature. \citet{bao-etal-2022-aeg} first proposed ArgEssay, a large-scale standard dataset that collected 11,000 high-quality argumentative essays. In AEG-related tasks, such as argument generation, we could also witness great attempts to utilize text-planning. \citet{hua-etal-2021-dyploc} used concepts and knowledge relevant to viewpoints as planning to increase the persuasiveness of argument generation. \citet{hu-etal-2022-planet} amplified Transformer decoder with latent representations and maintained sentence-level text-planning based on word packets, succeeding in generating more content-rich coherent text. 

However, all these existing methods failed in generating the structure of argumentative texts. Moreover, current planning methods often rely on unreadable hidden variables or a series of keywords, which is lack of interpretation for language models.

\subsection{Opinion Text Generation}
Opinion generation produces arguments for or against a given topic. Unlike the AEG task, generating a number of major claims is enough for opinion generation task and it does not require a full essay with detailed argumentative process. 

With the development of large-scale pre-trained generative language models, recent research has emphasized the direct generation of arguments using generative methods. \citet{gretz-etal-2020-workweek} devised an opinion generation pipeline using GPT-2. \citet{schiller-etal-2021-aspect} proposed a controlled opinion generation model capable of producing sentence-level arguments based on given topics, positions, and aspects. \citet{al-khatib-etal-2021-employing} developed three argumentative knowledge graphs and extracted knowledge from that to formulate prompts for training end-to-end opinion generation models. \citet{hu2023americano} proposed a agent interaction framework for opinion and counter-arguments generation using ChatGPT.

The above approaches mainly focused on generating single and brief arguments, without considering the generation of coherent long argumentative texts encompassing multiple arguments.

\begin{figure*}[!t]
\centering
\includegraphics[width=0.95\textwidth]{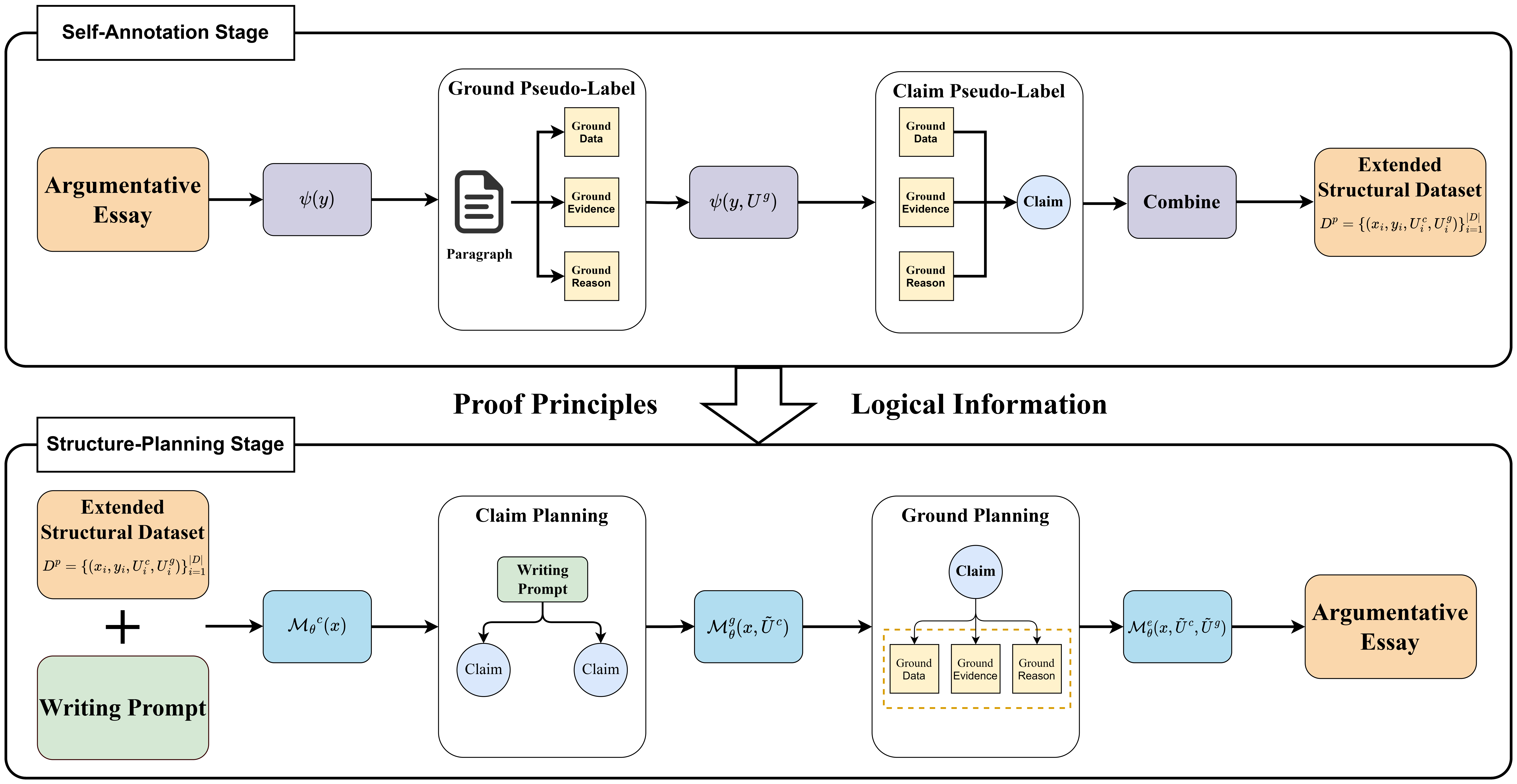}
\caption{The full flow chart of PESA. The upper figure shows the Proof-Enhancement process of generating text-planning from writing prompt and finally generating argumentative text, while the lower figure shows the Self-Annotation process of gradually building pseudo-labels for Proof-Enhancement training from ground truth using GPT-4. Detailed Proof-Enhancement samples are given in and Appendix \ref{logical_structure} and Appendix \ref{text-planning-design-app} .}
\label{model-graph}
\end{figure*}

\section{Method}

\subsection{Overview}

The AEG task can be formally defined as follows: given a writing prompt $X=\{w_1^x,w_2^x,...,w_n^x\}$, generate coherent and persuasive argumentative text $Y=\{w_1^y,w_2^y,...,w_m^y\}$ associated with the writing prompt.

Drawing inspiration from the theory of Toulmin Argumentation Model, we propose a unified framework that brings hierarchical proof principles and logical information i.e., claims and grounds, to AEG task, named PESA.
As shown in Figure \ref{model-graph}, PESA is consists of two approaches: (i) Self-Annotation and (ii) Proof-Enhancement.

The Self-Annotation approach creates a model $f$, that inputs a writing prompt $X$ and its ground truth $Y$, and outputs the fine-grained logical pseudo labels $U^h$, such that, $f(X, Y) = U^h$. 
Next the Proof-Enhancement approach employs a model $g$ that uses the pseudo labels $U^h$ and the writing prompt $X$ to generate argumentative text , that is $g(X, U^h) = Y$.




\subsection{Proof-Enhancement}
Proof-Enhancement abstract Toulmin Argumentation Model into two-tier tree to finely organize claim proof flow and detail grounds. The first level generates a set of major claims to dictate the overall logical structure. The second level generates data, evidence, and other supplemental writing materials to guide sentence-level writing.

Proof-Enhancement approach introduces structural information using two levels of tree text-planning. Specifically, this planning contains two components: (i) \emph{Claims Planning}: a list of major claims based on the essay’s topic, and each of them should build a complete logical chain to argue for a particular topic or express a specific point of view. (ii) \emph{Grounds Planning}: a list of supporting evidence or the process of logical reasoning to justify each claim respectively. Finally, the generative model expands the Proof-Enhancement into a complete argumentative essay. 

These two-tier planning are formally notated as $U^{c}$ and $U^{g}$. Notably, unlike most of the previous work, we employ $U^{c}$ and $U^{g}$ constructed in natural language similar to a human outline instead of using unreadable hidden variables or embedding. This is because utilizing methods like self-instruction learning that allows models to understand natural, readable planning, can be seen as a more user-friendly approach for generative LLM.

More specifically, the first step is to train the model $\mathcal{M}_{\theta}^{c}$, this model aims to generate the major claims for each paragraph, considering their relevance to the given topic and the logical structure, such as juxtaposition and progression among these claims, which shape the overall logical flow of the full text. 
\begin{align}
\tilde{U}^{c} &= \mathcal{M_{\theta}}^{c}(x) .
\end{align}
After that, the next model $\mathcal{M}_{\theta}^{g}$ is trained. This model needs to generate evidence and data for each major claim in form of brief paragraph, considering proper logical reasoning to support the claim accordingly, which consist the intra-paragraph logical structure. 
\begin{align}
\tilde{U}^{g} &= \mathcal{M}_{\theta}^{g}(x, \tilde{U}^{c})  .
\end{align}
Finally, the generate model $\mathcal{M}_{\theta}^{e}$ expands claim planning and ground planning to complete the AEG task. 
\begin{align}
\tilde{y} &= \mathcal{M}_{\theta}^{e}(x, \tilde{U}^{c}, \tilde{U}^{g}) .
\end{align}
Figure \ref{model-graph} illustrates detailed process diagram of this method, while figure in Appendix \ref{text-planning-design-app} provides an in-depth overview of the Proof-Enhancement design. Through this methodology, a Toulmin Model structure is explicitly introduced into the generation process, enabling the model to generate human-like logical structures that better fulfill the requirements of the AEG task.

\subsection{Self-Annotation}
Self-Annotation data augmentation method resembles the condensation process of discursive text examples, constructing text-planning ground truth step by step from the argumentative text paradigm. Tree logical structure is attaining through a layer-by-layer summary of the text, yielding both abstract claims and specific grounds. 

The process of Self-Annotation acts as the inverse process of the Proof-Enhancement. Drawing inspiration from the hierarchical summary technique proposed by \citet{wu2021recursively}, we construct claim and ground planning pseudo-labels using logical information naturally contained in multi-layer summaries. For scalability reasons, we use LLM automatic annotation instead of human annotation to generate pseudo-labels for Proof-Enhancement training. Manual evaluation shows that our proposed Self-Annotation method has similar accuracy and high consistency with human labeling results.

For each writing prompt $x \in X$ and argumentative text ground truth $y \in Y$, firstly, we use GPT-4 to generate summaries for each paragraph of $y$. These summaries are instructed to contain complete logical structure and grounds such as examples and logical reasoning sentences, and serve as the ground planning $U^{g}$. Then these summaries are used as a foundation to extract the paragraph major claims, which could constitute the claim planning $U^{c}$. 
\begin{align}
U^{g} &= \psi (y) ,\\
U^{c} &= \psi (y, U^{g}) ,
\end{align}
where $\psi$ is a function to summaries $y$ layer-by-layer using LLM to extract logical and proof information from the given text. In this way, we extend the labeled dataset $D$ into a pseudo-labeled dataset $D^p=\{(x_i, y_{i}, U^{c}_i, U^{g}_i)\}_{i=1}^{|D|}$. Intuitively, logical information can bring additional supervision signals for training Proof-Enhancement.



We illustrate this pair of processes in Figure \ref{model-graph}, it is clear that the Self-Annotation process is similar to long text compression. Several studies have shown that LLMs such as ChatGPT and GPT-4, are good data annotators and generators because they could produce data whose quality is comparable to those manually created by humans. As a result, the quality of the generated text-planning using this approach can be guaranteed.

\subsection{Training and Inference}
PESA framework is designed to train three generative models: 
(i) a model $\mathcal{M}_{\theta}^c$ aims to generate claims planning
(ii) a model $\mathcal{M}_{\theta}^g$ aims to generate ground planning
and (iii) a model $\mathcal{M}_{\theta}^e$ aims to generate the whole argumentative essay.
The above three models are individually trained with the following loss functions:
\begin{align}
    \mathcal{L}_{c} &= -\sum_{t=1}^{|X|}{\log P(U_t^{c}|X_t)}  ,\\
    \mathcal{L}_{g} &= -\sum_{t=1}^{|X|}{\log P(U_t^{g}|U^{c},X_t)} ,\\
    \mathcal{L}_{e} &= -\sum_{t=1}^{|X|}{\log P(Y_t|{U}^{g}{,U}^{c},X_t)} ,
\end{align}
where $X$, $U^{c}$, $U^{g}$, and $Y$ represent writing prompt, claims planning, ground planning, and argumentative essay, respectively. 
We introduce an algorithm using pseudo-label data and Proof-Enhancement, as shown in Algorithm~\ref{PESA:training}.

As for the inference process, we generate $U^{c}$, $U^{g}$ and $Y$ sequentially using a form of pipeline, and all generation steps are done using an autoregressive large language model.

\section{Experimental Setting}

\begin{algorithm}[t!]
    \caption{Training PESA framework}
    \label{PESA:training}
    \SetKwData{Left}{left}\SetKwData{This}{this}\SetKwData{Up}{up} \SetKwFunction{Union}{Union}\SetKwFunction{FindCompress}{FindCompress} \SetKwInOut{Input}{input}\SetKwInOut{Output}{output}
    \Input{Writing prompt $X$, essay ground truth $Y$, model $\mathcal{M}^{c}_{\theta}$, $\mathcal{M}^{g}_{\theta}$ and $\mathcal{M}^{e}_{\theta}$}
    \Output{Updated model $\mathcal{M}^{c}_{\theta^{'}}$, $\mathcal{M}^{g}_{\theta^{'}}$ and $\mathcal{M}^{e}_{\theta^{'}}$}
    \BlankLine
    \For{each batch in epoch}{
        \For{each sample $(x, y)$ in batch} {
            $U^{g} \gets \psi (y)$;

            $U^{c} \gets \psi (y, U^{g})$;
            
            $\tilde{U}^{c} \gets \mathcal{M_{\theta}}^{c}(x)$;
            
            $\tilde{U}^{g} \gets \mathcal{M}_{\theta}^{g}(x, \tilde{U}^{c})$;
            
            $\tilde{y} \gets \mathcal{M}_{\theta}^{e}(x, \tilde{U}^{c}, \tilde{U}^{g})$;
            
            $\mathcal{L}_{c} \gets f_{c}(\tilde{U}^{c}, U^{c})$;

            $\mathcal{L}_{g} \gets f_{g}(\tilde{U}^{g}, U^{g})$;

            $\mathcal{L}_{e} \gets f_{e}(\tilde{y}, y)$;
        }
        Minimize loss $\mathcal{L}_{c}$, $\mathcal{L}_{g}$, $\mathcal{L}_{e}$ and update parameters $\theta \gets \theta'$;
    }
    Back to line 1 to iterate lines 1--11 until the maximum epoch.
\end{algorithm}

\subsection{Dataset}
We evaluated our model on the ArgEssay\citep{bao-etal-2022-aeg}, the largest standardized assessment dataset currently available for the AEG task. This dataset comprises 11,000 high-quality argumentative essays and corresponding writing topics sourced from various international standardized English writing tests (e.g., IELTS and TOEFL). It covers a range of common controversial topics such as technological advances, educational approaches, and environmental issues. The final dataset consists of 11,282 writing topic-argumentative text pairs, of which 9,277 pairs are allocated to the training set, 1,002 pairs to the validation set, and 1,003 pairs to the test set.

\subsection{Baselines}

\begin{table*}[!t]
\centering
\resizebox{\textwidth}{!}{
\begin{tabular}{{l}*{8}{c}}
  \toprule
  Method & Relevance & \makecell{Validity of \\ Reasoning} & \makecell{Credibility of \\ Evidence} & \makecell{Language and \\ Rhetoric} & \makecell{Overall \\ Persuasiveness}  \\
  \midrule
  DD-KW & 58.80 & 52.25 & 25.00 & 54.13 & 52.26 \\
  DD-Rel & 60.73 & 54.45 & 25.93 & 55.48 & 53.66 \\
  LLaMA2-base & 90.40 & 80.26 & 45.92 & 72.28 & 70.54 \\
  LLaMA2-CoT & 90.42 & 82.14 & 46.20 & 73.35 & 72.12 \\
  \textbf{Ours} & \textbf{90.75} & \textbf{84.64} & \textbf{49.20} & \textbf{75.52} & \textbf{75.04} \\
  \emph{-w/o claim planning} & 89.23 & 82.35 & 48.02 & 73.12 & 72.42  \\
  \emph{-w/o ground planning} & 90.00 & 83.33 & 46.92 & 72.34 & 73.19  \\ 
  \bottomrule
\end{tabular}
}
\caption{\label{automatic-exp}
The results of comparison of baselines on automatic metrics. \textbf{Bold} numbers denote the best performance among all methods on each dataset.
}
\end{table*}

We compared our proposed approach with several strong baseline models to showcase its effectiveness. Given the absence of typical LLM-based approaches in the AEG task for fair comparison, we established several LLM-based strong baselines and conducted comprehensive comparisons.

\textbf{DD-KW.} The text-planning method proposed by \citet{bao-etal-2022-aeg} . This method uses a series of keyword sets as planning, and designs a dual decoder structure based on BART, one is a planning decoder (PD) to generate text-planning, and the other is a writing decoder (WD) finally generates argumentative texts.

\textbf{DD-Rel.} Another text-planning method proposed by \citet{bao-etal-2022-aeg} using same framework, and changes the text-planning from keywords to the relational triples drawn from open source knowledge graph.

\textbf{LLaMA-base.} \citet{touvron2023LLaMA} presents a collection of baseline language models from 7B to 65B parameters LLaMA. LLaMA-13B achieve SOTA results on various benchmarks, which is a very strong baseline model especially for text generation tasks. We fine-tuned LLaMA2-13B-chat as an end-to-end generation model for AEG task named LLaMA-base. 

\textbf{LLaMA-CoT.} \citet{wei2022chain} presents a chain of thought prompt that significantly improves the ability of large language models without training. We design a CoT prompts on the LLaMA2 base model to fit the AEG task named LLaMA2-CoT.

\subsection{Evaluation Metrics}

Automatically evaluating open-domain text generation text is a challenging task\citep{celikyilmaz2021evaluation}. For the AEG task, evaluating the quality of generated text solely based on automatic metrics such as BLEU can be insufficient, as there can be multiple valid argumentative essays in the ground truth. To solve this problem, some works in recent years have utilized GPT-4 for ground-truth-free evaluation, which significantly enhance the consistency of automatic evaluation metrics with manual evaluation\citep{hu2023americano,liu2023geval,fu2023gptscore}. Inspired by these work, we propose a GPT-4-based AEG evaluation approach. 

\textbf{Automatic evaluation.}
Based on the systems theory of argumentation\citep{van2004systematic}, we evaluate the persuasive of argumentative essay from the following five aspects: \emph{Relevance}, \emph{Validity of the Reasoning}, \emph{Credibility of Evidence}, \emph{Language and Rhetoric} and \emph{Overall Persuasiveness}. We design prompts for each evaluation aspect with specific task instructions and a comprehensive list of detailed criteria. To reduce the variance of the evaluation results, we prompt model to give a detailed explanation for evaluation results before the final prediction of score. Detailed prompts for each evaluation aspect will be given in Appendix \ref{prompts}.

\textbf{Human evaluation.}
We employed three skilled English speakers with NLP background to perform manual evaluation of the generated results. We set up the following two evaluation tasks: 
(i) \emph{Scoring the generated text.} Same as GPT-4 based evaluation, score the generated results in terms of five aspects with the same criteria as the prompt used in GPT-4. 
(ii) \emph{Compare the results with baselines.} The annotator will compare the outputs of our proposed method with all the baselines and give an overall evaluation of \emph{Win}, \emph{Loss} and \emph{Tie} for each test sample pair.

\subsection{Implementation Details}
We finetune LLaMA2-13B-chat with all parameters with the help of huggingface and DeepSpeed. AdamW optimizer is adopted for optimization, and initial learning rates are set to 1e-5 with linear descent schedule. We train the model 5 epochs. The batch size per-device is set to 8. All experiments are conducted with NVIDIA Tesla A100 GPU.

\section{Results and Analysis}

\begin{table*}[!t]
\centering
\resizebox{\textwidth}{!}{
\begin{tabular}{{l}*{5}{c}}
  \toprule
  Method & Relevance & \makecell{Validity of \\ Reasoning} & \makecell{Credibility of \\ Evidence} & \makecell{Language and \\ Rhetoric} & \makecell{Overall \\ Persuasiveness}  \\
  \midrule
  DD-KW & 3.82 & 2.86 & 3.30 & 3.04 & 3.00 \\
  DD-Rel & 3.72 & 2.90 & 3.34 & 3.08 & 3.00 \\
  LLaMA2-base & 4.92 & 4.34 & 4.46 & 4.46 & 4.50 \\
  LLaMA2-CoT & 4.94 & 4.46 & 4.68 & 4.58 & 4.62 \\
  ChatGPT & \textbf{4.97} & 4.62 & \textbf{4.88} & \textbf{4.70} & \textbf{4.82} \\
  \textbf{Ours} & 4.94 & \textbf{4.68} & 4.76 & 4.60 & 4.76 \\
  \bottomrule
\end{tabular}
}
\caption{\label{human-exp}
The results of comparison of baselines on human evaluation metrics. \textbf{Bold} numbers denote the best performance among all methods on each dataset.
}
\end{table*}

\subsection{Main Results}

\textbf{Performance on automatic evaluations.} As shown in Table \ref{automatic-exp}, experimental results demonstrate the promising potential of the PESA framework. For both Relevance, Validity of the Reasoning, Credibility of Evidence, Language and Rhetoric and Overall Persuasiveness, our method demonstrates varying degrees of improvement. Notably, our framework has exhibited more significant improvements in the Validity of the Reasoning and Credibility of Evidence metric, which assess logical structure and correctness. These improvements indicates a more reliable logical structure and persuasive argumentative essay. Specifically, we observe 0.35 points improvement on Relevance metric, 4.38 points improvement on Validity of the Reasoning, 3.28 points improvement on Credibility of Evidence metric, 3.24 points improvement on Language and Rhetoric metric and 4.50 points improvement on Overall Persuasiveness. Furthermore, experimental results show that the claim and ground planning effects the generation results in different ways. Claim planning contributes more to the Validity of Reasoning metric, while the effects of ground planning are more focused on the Credibility of Evidence metric. This distinction arises because claim planning controls the essay's overarching logical structure, thereby significantly influencing reasoning generation. In contrast, ground planning manages the detailed substantiation process of the claim, affecting the generation of evidence. These finding indicates the efficacy of the PESA method, confirming its alignment with our motivation.

\begin{figure}[t]
\centering
\includegraphics[width=0.48\textwidth]{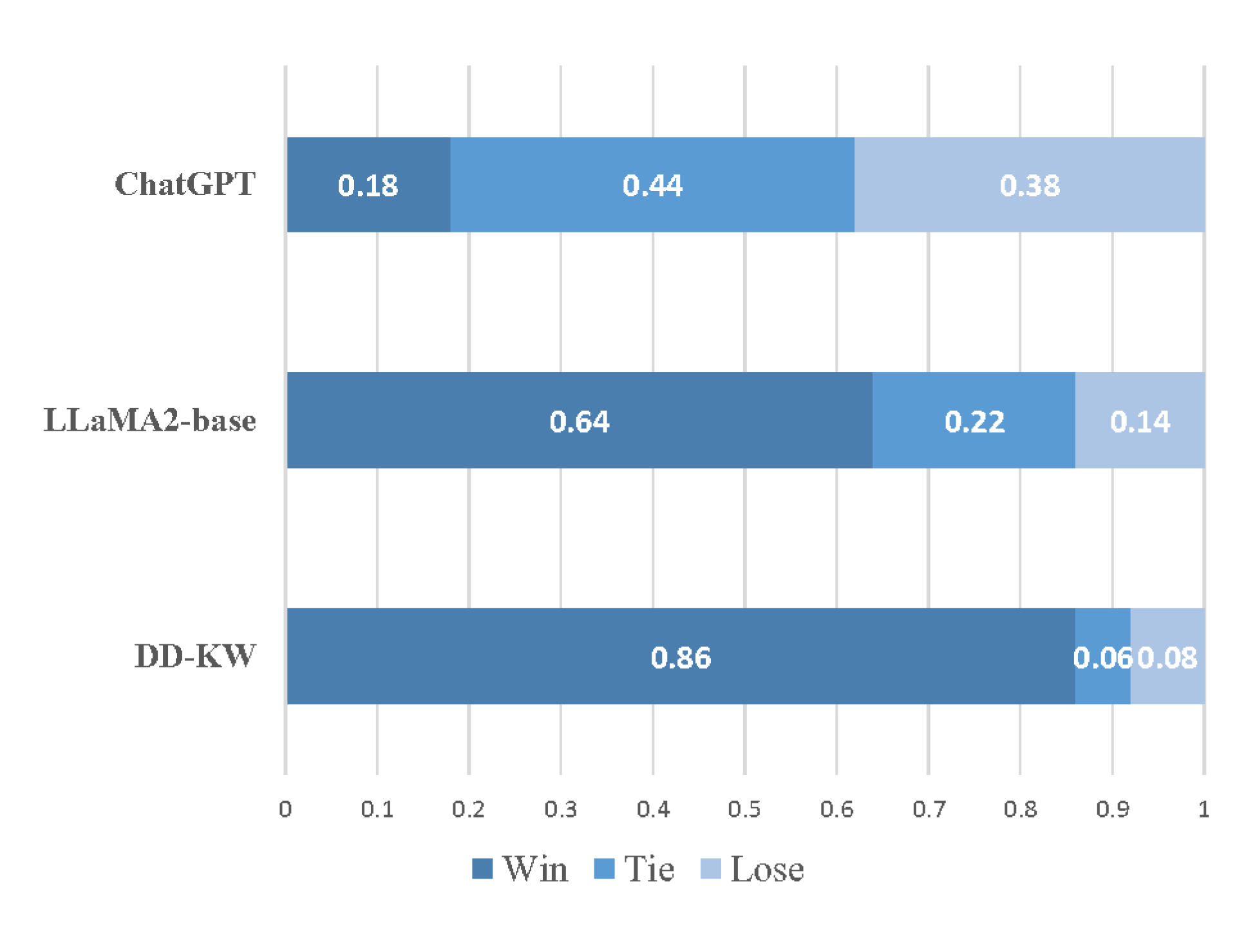}
\caption{PESA compared to other baselines. Human raters compared different model generations and and scored them accordingly.}
\label{compared_results}
\end{figure}

\textbf{Performance on human evaluation.} The human evaluation metrics are presented in Table \ref{human-exp} and Figure \ref{compared_results}. When compared to the state-of-the-art model with the same parameter size like LLaMA2-base, our PESA method demonstrates superior performance across all five evaluated aspects. This suggests that our proposed approach effectively directs the model towards generating diverse and high-quality texts. In comparison to larger language models, such as ChatGPT, the PESA method achieves comparable outcomes across all five aspects and surpasses ChatGPT in Validity of Reasoning. The direct comparison of our method with strong baseline models, as depicted in Figure \ref{compared_results}, reveals that our model exceeds the current state-of-the-art models, DD-KW and LLaMA, by 64\% and 86\%, respectively, and is not inferior to ChatGPT by 62\%. Taken together, these experimental and human evaluation results indicate that the PESA method not only significantly outperforms models of similar parameter size but also competes favorably with models of larger parameter sizes.

\subsection{Ablation Study}

\textbf{Effectiveness of Proof-Enhancement Method.} We evaluate the performance of Proof-Enhancement without the claim planning and without the ground planning. As shown in Table \ref{automatic-exp}, experimental results show that the performance of the PESA framework decreases when either the claim planning or ground planning is removed, which indicates that both of them play a positive role. Notably, claim planning and ground planning have different impacts on the results. Claim planning exerts the greatest influence on the Validity of Reasoning metric, surpassing ground planning by 1.02 points. Conversely, ground planning has the largest effect on the Language and Rhetoric metric, exceeding claim planning by 1.1 points. This distinction reveals that within Proof-Enhancement, claim planning guides more macro-level connections, while ground planning affects more granular aspects. Such findings align with the intended design motivations of Proof-Enhancement and demonstrate its effectiveness.

\textbf{Effectiveness of Self-Annotation Data Augmentation.} Further, in Table \ref{data-exp}, we analyze the effectiveness of different base models on the Self-Annotation method by manual evaluation. The results show that GPT-4 performs well in both relevance and quality of the generated plannings. Meanwhile, the selection of LLaMA2-13B caused only about 15\% performance loss in data construction, so our method also has good compatibility with smaller models.

\subsection{Case Study}

\begin{table}[]
\centering
\small
\begin{tabular}{{l}*{2}{c}}
  \toprule
  Model & Relevance & Quality  \\
  \midrule
  LLaMA2-13B & 4.90 & 3.75  \\
  ChatGPT & 4.92 & 4.15  \\
  GPT-4 & \textbf{4.95} & \textbf{4.33}  \\
  \bottomrule
\end{tabular}
\caption{\label{data-exp}
Comparison of the effects of different models on the construction of fine-grained training Data. The human annotator evaluated 50 samples in terms of both relevance and quality out of 5 points. \textbf{Bold} numbers denote the best performance among all methods on each dataset.
}
\end{table}

We give an example in the Appendix \ref{example-outputs} that includes all baselines and our model outputs. In this example, the Writing Prompt is ``In countries with a high rate of unemployment, numerous pupils should do offered only predominant education. There is no point in offering secondary education to those who have no hope of finding a job. To what extent do you agree or disagree with this statement''. It can be seen that our method and ChatGPT provide a clear and complete global logic chain, and also have good performance in logical coherence within paragraphs. 

In contrast, the two claims given by the LLaMA2-base are not deeply related to unemployment in Writing Prompt, and the internal logic is not smooth with no examples or logical reasoning to prove its own opinions. The DD-KW method generated an article that overall conforms to the topic, but the logical structure is very confusing. In the first paragraph, DD-KW believes that ``the government should not offer tertiary education to those who are unemployed'' is wrong, but the following two major claims turn to support this view. The DD-Rel method is similar to the DD-KW method and produces serious logical errors. For example, DD-Rel believes that ``offering secondary education to pupils who have no hope of finding employment is not beneficial'', but in the following two paragraphs, one introduces university education at length, and the other is more inclined to support this view.

\section{Conclusion}

In this paper, we present a unified two-stage framework (PESA) for AEG with a focus on proof and logical correctness. The framework is composed of two components: The Self-Annotation method leverages a large language model to progressively extract logical information from AEG data, obtaining high-quality training data; the Proof-Enhancement method enhances the persuasive power of generative results by explicitly incorporating proof principles into the planning process, enabling the language model to produce argumentative texts in line with the Toulmin Argumentation Model. Automatic and manual evaluation metrics show that our method could produce fluent, coherent, and persuasive argumentative text with a significant improvement in the logical structure.

\section*{Limitation}
Despite the excellent results of our model on the AEG task, a potential limitation exists in our model due to the absence of explicit modeling for causal relationships among major claims, topics, and writing materials within paragraphs. In future studies, we plan to use causal modeling approaches to target logical structures in argumentative texts for more in-depth control.

\section*{Ethical Consideration}
The ethical risks of our proposed methods and models are low. This is because we used a publicly published dataset that has passed ethical review and does not contain sensitive or private content. Moreover, the open source models we use such as LLaMA have undergone strict security training, and the output content complies with ethical standards.

\section*{Acknowledgement}
This research was supported by the National Key Research and Development Program (No. 2022YFF0902100) and National Natural Science Foundation of China (No. 62076081 and No. 61936010) and Nature Scientific Foundation of Heilongjiang Province (YQ2021F006) and Du Xiaoman (Beijing) Science Technology Co., Ltd.

\bibliography{custom}

\newpage
\appendix

\section{Example of Argumentative Essay Logical Structure}
\label{logical_structure}
\begin{figure*}[t]
\centering
\includegraphics[width=0.95\textwidth]{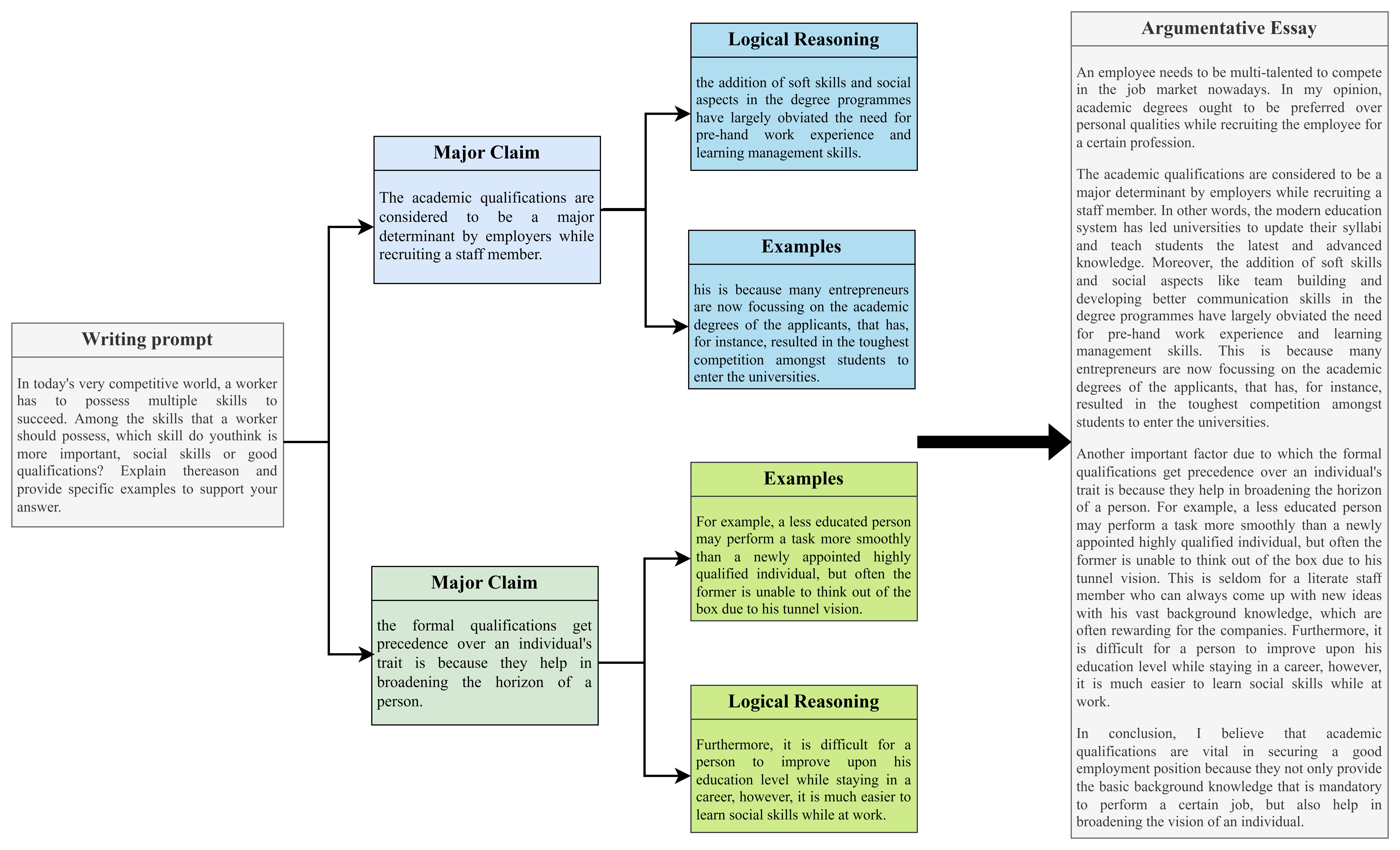}
\caption{Example of the logical structure in human-authored argumentative text. The leftmost writing prompt extends two Major claims, after which each Major claim expands into several grounds or evidence.}
\label{hierarchical structure}
\end{figure*}

An example of logical structure that exists in real argumentative essays is given in Figure \ref{hierarchical structure}.

\section{Detailed Design Of Proof-Enhancement}
\label{text-planning-design-app}
\begin{figure*}[t]
\centering
\includegraphics[width=0.95\textwidth]{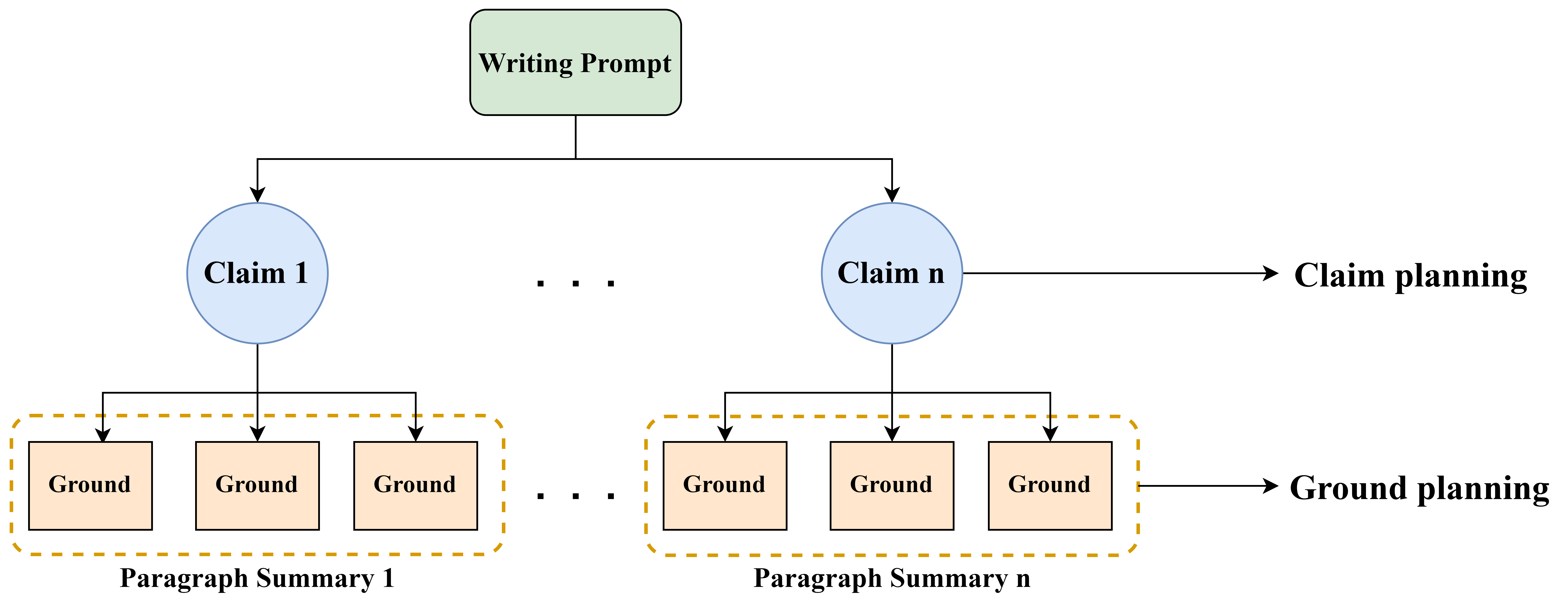}
\caption{The specific design of Proof-Enhancementg. Two levels of text-planning are shown from top to bottom: the first level is the claim planning contains major claim, and the second level is the ground planning contains grounds, evidence and writing material.}
\label{text-planning-design}
\end{figure*}

The detailed design of the Proof-Enhancement is given in Figure \ref{text-planning-design}.

\section{Automatic Evaluation Prompts}
\label{prompts}

For GPT4-based automatic evaluation as described in section Experiments, we present the detailed prompts for each metric in Table \ref{Prompts}. The scoring range of the model is given after each prompt, and the model is required to give a brief explanation before scoring. This is to constrain the model output and reduce the randomness and variance.

\begin{table*}[]
\small
\centering
\resizebox{\textwidth}{!}{
\begin{tabular}{cc}
  \toprule
   Metric & Prompt \\
  \midrule
   Relevance & \multicolumn{1}{m{200pt}}{Rate the direct relevance of the above argumentative essay to the topic. The criterion is: all the claims in the argumentative essay should be related to the topic or the major claim of the essay. Rating score is out of 100. Please give a short reason for the rating first, followed by a direct score in the form of xx/100.} \\ 
   \midrule
   Validity of the Reasoning & \multicolumn{1}{m{200pt}}{Rate the clarity and coherence of the logic of the above argumentative essay. The criteria are: the main ideas presented in the essay should have logical relationships, including progression, concessions, etc., and be able to argue the major claim of the essay. Each main idea should be logically and consistently argued.  Rating score is out of 100. Please give a short reason for the rating first, followed by a direct score in the form of xx/100.} \\ \midrule
   Credibility of Evidence & \multicolumn{1}{m{200pt}}{Rate the credibility of the evidence used in the above argumentative essay. The criterion is: ideas should be supported by credible evidence or sources, and unsubstantiated ideas are not as persuasive as rebuttals based on corroborating evidence. Rating score is out of 100. Please give a short reason for the rating first, followed by a direct score in the form of xx/100.} \\ \midrule
   Language and Rhetoric & \multicolumn{1}{m{200pt}}{Rate the effectiveness of the language and rhetoric used in the above argumentative essay. The criterion is: argumentative essay should effectively use persuasive language and rhetorical techniques, such as appealing to logic or evidence, to convince the reader. Rating score is out of 100. Please give a short reason for the rating first, followed by a direct score in the form of xx/100.} \\ \midrule
   Overall Persuasiveness & \multicolumn{1}{m{200pt}}{Rate the overall persuasiveness of the above argumentative essay. The criterion is: the argumentative essay should effectively argue its point of view in a way that the reader is convinced and agrees with it. Please give a short reason for the rating first, followed by a direct score in the form of xx/100.} \\
  \bottomrule
\end{tabular}
}
\caption{\label{Prompts}
Prompts for all GPT4-based automatic metrics.
}
\end{table*}

\section{Example of Outputs}
\label{example-outputs}
We present additional examples with different model outputs in Table \ref{case study 1} and \ref{case study 2}.

\begin{table*}[]\tiny
\centering
\resizebox{\textwidth}{!}{
\begin{tabular}{p{250pt}}
  \toprule
  \textbf{Writing Prompt}: In countries with a high rate of unemployment, numerous pupils should do offered only predominant education. There is no point in offering secondary education to those who have no hope of finding a job. To what extent do you agree or disagree with this statement. \\
  \midrule
  \textbf{DD-KW}:These days, a large number of the students are unable to finish their studies in the secondary schools due to the reasons. Some argue that the government should not offer tertiary education to those who are unemployed. I completely disagree with this view and in this essay, I will discuss my own opinion in detail. First and foremost, offering secondary education to the students Suggestionin the primary schools is not beneficial to all of the pupils. If the primary school does not have enough resources to provide higher education, it will not be beneficial for the students. For example, if the secondary school does offer only primary school, the students who are unable find a job will not tend to study there. In addition, the primary institution will not have adequate resources to train the students with the right skills. Moreover, the secondary education will not provide the job seekers with sufficient opportunities to work. The children who are not able to find a well-paid job will be unable to afford the tuition fees. As a result, they will not get the necessary skills and knowledge to continue their studies. For instance, In the USA, the majority of the school leavers are unable as they are not capable of learning the subjects required to become a skilled worker. Therefore, it is apparent that the secondary level will not benefit the students as they do not have sufficient resources to study. In conclusion, offering primary education has few advantages. The government should provide a job opportunity to the pupils who are unready to study the subjects. \\
  \textbf{DD-Rel}: In some countries, many pupils are keen to get only predominant education. However, I believe that offering secondary education to pupils who have no hope of finding employment is not beneficial. I will explain why many students are keen on focusing on their secondary education and the reasons why I agree with this view. Firstly, focusing on secondary education is a waste of money. For example, the price of a university course is much higher than another school course. For this reason, some pupils prefer to study a school subject that is related to their interest. As a consequence, they are willing to pay more money for a good university. Moreover, they feel that studying the subject that they really like will help them to get a job. It means that if students study only secondary education, they will not get a good job in the future. Secondly, focusing only on primary education can be harmful to the students. There are many subjects which are useless in the current market. For instance, English is considered to be less important than science. Many students who have a talent in English tend to choose a programming program that is not related to science, which is not useful for their job. This means that they will be more likely to be unemployed. In conclusion, in my opinion, it is not an effective way to offer a secondary education for those who do not have any hope of getting a job right away. Therefore, focusing more on secondary studies can be a better way to encourage students to study. \\
  \textbf{LLaMA2-base}:In the developing countries, the unemployment is a major problem. Some people think that the students should be provided with the vocational training only, as the secondary education is useless for them. I completely disagree with this statement. Firstly, the education is the basic right of every human being. The government should provide the education to all its citizens. The unemployment is not the fault of the students. They should be provided with the education, so that they can also serve the nation in the future. Moreover, the education is the only way to reduce the poverty and the crime rate in the society. Secondly, the education is not only the right, but it is also the responsibility of the government. The government should provide the education to all its citizens. The government should not discriminate between the rich and the poor, the literate and the illiterate. The government should provide the education to all its citizens. The education is the only way to reduce the poverty and the crime rate in the society. Thirdly, the education is the basic requirement of the human being. The education is the only way to reduce the poverty and the crime rate in the society. The education is the only way to serve the nation. The education is the only way to get the job. The education is the only way to get the respect in the society. To conclude, the education is the basic right of every human being. The education is the only way to reduce the poverty and the crime rate in the society. The education is the only way to serve the nation. The education is the only way to get the job. The education is the only way to get the respect in the society. \\
  \bottomrule
\end{tabular}
}
\caption{\label{case study 1}
Output example of different baselines and ours.
}
\end{table*}

\begin{table*}[]\tiny
\centering
\resizebox{\textwidth}{!}{
\begin{tabular}{p{250pt}}
  \toprule
  \textbf{Writing Prompt}: In countries with a high rate of unemployment, numerous pupils should do offered only predominant education. There is no point in offering secondary education to those who have no hope of finding a job. To what extent do you agree or disagree with this statement. \\
  \midrule
  \textbf{ChatGPT}:In some countries, many pupils are keen to get only predominant education. However, I believe that offering secondary education to pupils who have no hope of finding employment is not beneficial. I will explain why many students are keen on focusing on their secondary education and the reasons why I agree with this view. Firstly, focusing on secondary education is a waste of money. For example, the price of a university course is much higher than another school course. For this reason, some pupils prefer to study a school subject that is related to their interest. As a consequence, they are willing to pay more money for a good university. Moreover, they feel that studying the subject that they really like will help them to get a job. It means that if students study only secondary education, they will not get a good job in the future. Secondly, focusing only on primary education can be harmful to the students. There are many subjects which are useless in the current market. For instance, English is considered to be less important than science. Many students who have a talent in English tend to choose a programming program that is not related to science, which is not useful for their job. This means that they will be more likely to be unemployed. In conclusion, in my opinion, it is not an effective way to offer a secondary education for those who do not have any hope of getting a job right away. Therefore, focusing more on secondary studies can be a better way to encourage students to study. \\
  \textbf{Ours}: In many countries, the unemployment rate is increasing at an alarming rate. Some people believe that students from poor families should only be offered predominant education and there is no point in offering secondary education to them as they will not be able to find a job in the future. I completely disagree with this statement and I will discuss why I disagree with this statement in this essay. Firstly, the main reason why I disagree with this statement is that the secondary education is the fundamental step for higher education. This education provides the necessary knowledge and skills to the students so that they can pursue higher studies. For example, in most of the countries, higher education is not accessible for the students who do not have secondary education. Therefore, secondary education is a pre-requisite for higher education. Secondly, secondary education is the basic right of every citizen. Every child has the right to education irrespective of their financial background and future job prospects. It is the responsibility of the government to ensure that every citizen has access to this basic right. For instance, in India, the right to education act of 2009 was passed to ensure that every child between the age group of 6 to 14 has a fundamental right to education. In conclusion, I strongly disagree with the statement that only predominant education should be offered to the students in countries with high unemployment rate. I believe that every child has a right to education and it is the responsibility of the government to ensure that every child has access to this right. \\
  \bottomrule
\end{tabular}
}
\caption{\label{case study 2}
Output example of different baselines and ours.
}
\end{table*}

\end{document}